\documentclass[conference]{IEEEtran}
\IEEEoverridecommandlockouts
\usepackage{cite}
\usepackage{amsmath,amssymb,amsfonts}
\usepackage{algorithmic}
\usepackage{graphicx}
\usepackage{textcomp}
\usepackage{xcolor}

\usepackage{mathptmx} 
\usepackage{times} 
\usepackage{amsmath} 
\usepackage{amssymb}  
\usepackage{array}
\usepackage{caption}
\usepackage{subcaption}
\usepackage{balance}
\usepackage{float}
\usepackage{url}
\usepackage{multirow}
\usepackage[export]{adjustbox}
\usepackage{footnote}
\usepackage{booktabs}
\usepackage{multirow}
\usepackage{enumitem}
\usepackage{soul}
\setlist{nosep, leftmargin=14pt}
\usepackage{makecell}
\usepackage{mwe} 

\def\BibTeX{{\rm B\kern-.05em{\sc i\kern-.025em b}\kern-.08em
    T\kern-.1667em\lower.7ex\hbox{E}\kern-.125emX}}
    
\begin{document}

\title{A Novel Transparency Strategy-based Data Augmentation Approach for BI-RADS Classification of Mammograms }

\author{

\IEEEauthorblockN{Sam B. Tran$^{\dag}$}
\IEEEauthorblockA{\textit{Vingroup Big Data Institute} \\
\textit{10000, Hanoi, Vietnam} \\ \textcolor{blue}{samsamtranbao@gmail.com}
} 

\and

\IEEEauthorblockN{Huyen T. X. Nguyen$^{\dag}$}
\IEEEauthorblockA{\textit{Vingroup Big Data Institute} \\
\textit{10000, Hanoi, Vietnam} \\ \textcolor{blue}{nguyenhuyen.csp.bka@gmail.com}
}

\and

\IEEEauthorblockN{Chi Phan$^{\dag}$}
\IEEEauthorblockA{\textit{VinUni-Illinois Smart Health Center,} \\
\textit{VinUniversity, 10000 Hanoi, Vietnam}\\ \textcolor{blue}{21chi.pth@vinuni.edu.vn}}
\and 

\IEEEauthorblockN{ \hspace*{3cm} Ha Q. Nguyen}
\IEEEauthorblockA{ \hspace*{3cm} \textit{Vingroup Big Data Institute}, \\
\textit{ \hspace*{3cm} VinBigData JSC,} \\
\textit{ \hspace*{3cm} 10000 Hanoi, Vietnam} \\ \hspace*{3cm} \textcolor{blue}{v.hanq3@vinbigdata.org}
}

\and

\IEEEauthorblockN{ \hspace*{1cm} Hieu H. Pham}

\IEEEauthorblockA{ \hspace*{1.5cm} \textit{VinUni-Illinois Smart Health Center,} \\
\textit{\hspace*{1.2cm} College of Engineering and Computer Science,} \\ \hspace*{1.5cm} VinUniversity, 10000 Hanoi, Vietnam \\  \hspace*{1.3cm} Correspondence: \textcolor{blue}{hieu.ph@vinuni.edu.vn}}

\and

\IEEEauthorblockN{ \hspace*{3cm} $^{\dag}$ These authors contribute equally to the work and share the first authorship.}

}

\maketitle

\begin{abstract}
Image augmentation techniques have been widely investigated to improve the performance of deep learning (DL) algorithms on mammography classification tasks. Recent methods have proved the efficiency of image augmentation on data deficiency or data imbalance issues. In this paper, we propose a novel transparency strategy to boost the Breast Imaging Reporting and Data System (BI-RADS) scores of mammogram classifiers. The proposed approach utilizes the Region of Interest (ROI) information to generate more high-risk training examples for breast cancer (BI-RADS 3, 4, 5) from original images. Our extensive experiments on three different datasets show that the proposed approach significantly improves the mammogram classification performance and surpasses a state-of-the-art data augmentation technique called CutMix. This study also highlights that our transparency method is more effective than other augmentation strategies for BI-RADS classification and can be widely applied to other computer vision tasks. 
\end{abstract}
\begin{IEEEkeywords}
Mammogram, deep learning, data augmentation, abnormality detection, BI-RADS classification.
\end{IEEEkeywords}

\section{Introduction}
\label{sec:intro}
Breast cancer has currently become the most common cancer, based on statistics from the International Agency for Research on Cancer (IARC) in December 2020~\cite{iarc}. The American Cancer Society (ACS) stated that the average hazard proportion of a woman in the United States developing breast cancer during her life was about 13$\%$~\cite{acs}. WHO estimated that 2.3 million women were diagnosed with breast cancer, and there were about 685,000 deaths worldwide in 2020~\cite{who}. The expert recommendation for women at high risk of breast cancer is to take diagnostic screening annually to detect cancer earlier and receive effective treatments beforehand. Mammography is a prevalent X-ray examination for breasts and is employed in computer-aided diagnosis (CADx) systems to assist radiologists in assessing breast cancer risk. In particular, the BI-RADS score is used as a risk evaluation and quality assurance tool that supplies a widely accepted lexicon and reporting schema for breast imaging~\cite{birad}. This standard consists of seven assessment levels: BI-RADS 0 (incomplete), BI-RADS 1 (negative), BI-RADS 2 (benign), BI-RADS 3 (probably benign), BI-RADS 4 (suspicious for malignancy), BI-RADS 5 (highly suggestive of malignancy), and BI-RADS 6 (known biopsy-proven malignancy). There are several recent DL-based studies which focus on BI-RADS classification~\cite{multiviewembc,biradsdnn} rather than benign/malignant classification. However, malignancy cases are often much less than benign cases, leading to data issues consisting of data deficiency and data imbalance. 

Recently, several data augmentation strategies have been proposed to resolve this problem and boost further training efficiency~\cite{reviewaug, cutmix, mixup, cutout, gan1}. There are some deep learning techniques for medical image synthesis, but mainly using generative adversarial networks (GANs) ~\cite{reviewaug, gan1}. The potential of GANs for image processing issues is enormous because they can be trained to mimic any dataset. However, several systematic reviews of GANs ~\cite{gan1, gan2} reveal that there are still many challenging issues in using this technique for medical imaging, including training instability, computational cost, and concerns regarding the quality of the generated samples. Whereas, some basic augmentation techniques such as cropping, filtering, flipping, and noise injection are simple to apply but still adequate for model performance improvement~\cite{reviewaug}. Two combination methods including~\cite{cutmix, mixup} could achieve remarkable results because a new image is generated by integrating some original images. Additionally, we observed that lesion areas in medical imaging, especially in mammograms, play an essential role, but have not been focused on by most current image augmentation techniques~\cite{augmentations}. This has motivated us to propose a new basic augmentation technique, called Transparency, for mammography BI-RADS classification. We conducted extensive experiments to show that the proposed approach has higher performance compared to a state-of-the-art data augmentation technique called CutMix~\cite{cutmix} in mammograms.

Our main contribution in this work is developing a transparency data augmentation technique that can generate new compelling images based on original images. A new image focuses on lesion areas without losing global image context by blurring the original image except for lesion areas. The new image would still have the same distribution as the original one and a deep focus on lesions. The proposed method is easy to apply to any medical dataset whose images have lesion bounding boxes and can be widely applied to other computer vision tasks. Experimental results indicate that our transparency technique outperforms the typical augmentation techniques by 2.7$\%$, 4.3$\%$, and 6.9$\%$  on the MIAS, our private dataset and VinDr-Mammo, respectively. It also surpasses the state-of-the-art CutMix~\cite{cutmix} on our private mammography dataset and VinDr-Mammo. The rest of the paper is organized as follows. The methodology is provided in Section~\ref{sec:methodology}. Our experiments are presented in Section~\ref{sec:ExperimentResult}. Finally, Section~\ref{sec:conclude} discusses the experimental results and concludes the paper.

\section{Methodology}
\label{sec:methodology}
\subsection{Preprocessing}
\label{ssec:subhead31}
One of the essential methods to improve the performance of machine learning models is data pre-processing, which is a group of different techniques to generate the most informative dataset for the training process. This section provides a brief detail of pre-processing data methods for our BI-RADS classification research. Four strategies Cropping, Flipping, CutMix, and Transparency will be discussed in this section.
\\ \subsubsection{Image cropping and flipping}
\label{sssec:subsub311}
\begin{figure}[t]
    \centering
    \includegraphics[width=\linewidth]{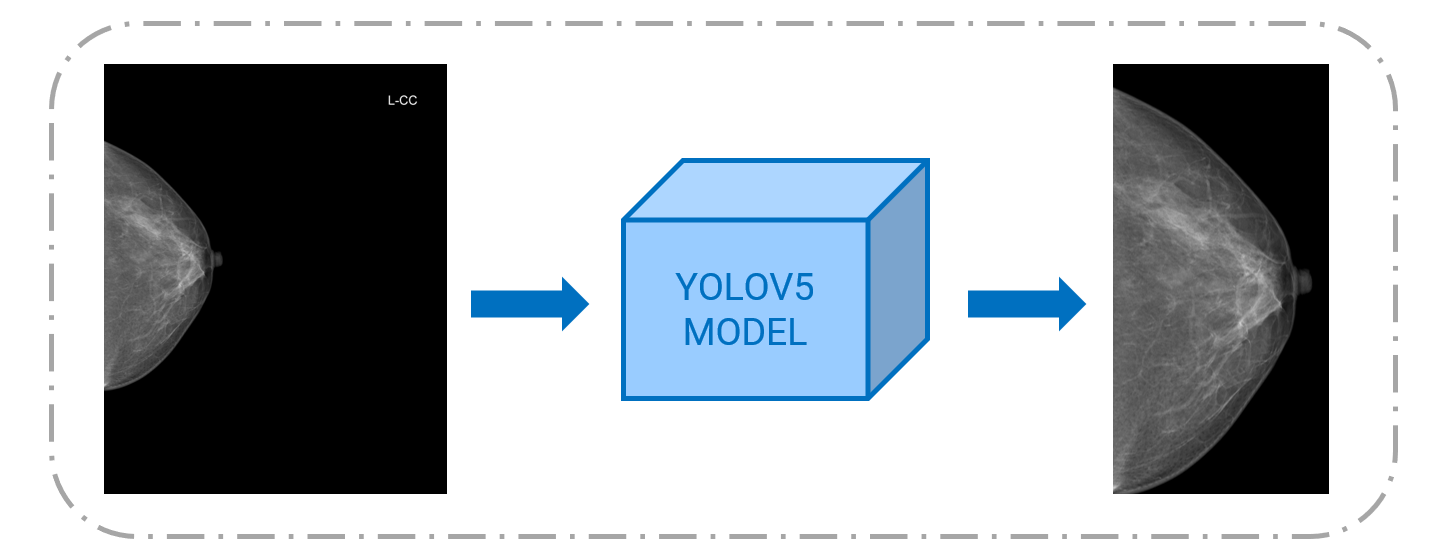}
    \caption{Image cropping by Yolov5~\cite{yolov5}.}
    \label{fig:yolov5}
  \end{figure}

One limitation of mammograms is that they contain a large black area without valuable information. Hence, we implement a breast detection model based on YOLOv5~\cite{yolov5} to crop the breast out of the original image. First, we used the labeling tool to label 2,000 images from the internal dataset. Then a dataset is built with 1,600 samples for training and 400 samples for validation. The medium YOLOv5~\cite{yolov5}  achieved the mAP of 0.995  for breast detection during testing. The cropping step can reduce input size before entering the DL model, avoiding wasting resources and shortening the computational time. An example of this process is illustrated in Figure~\ref{fig:yolov5}. We then flipped all the right mammograms along the vertical axis. By this way, we obtained a dataset where all the breast images have the same orientation.
\begin{figure*}
     \centering
     \begin{subfigure}[b]{0.48\textwidth}
         \centering
         \includegraphics[scale=.27]{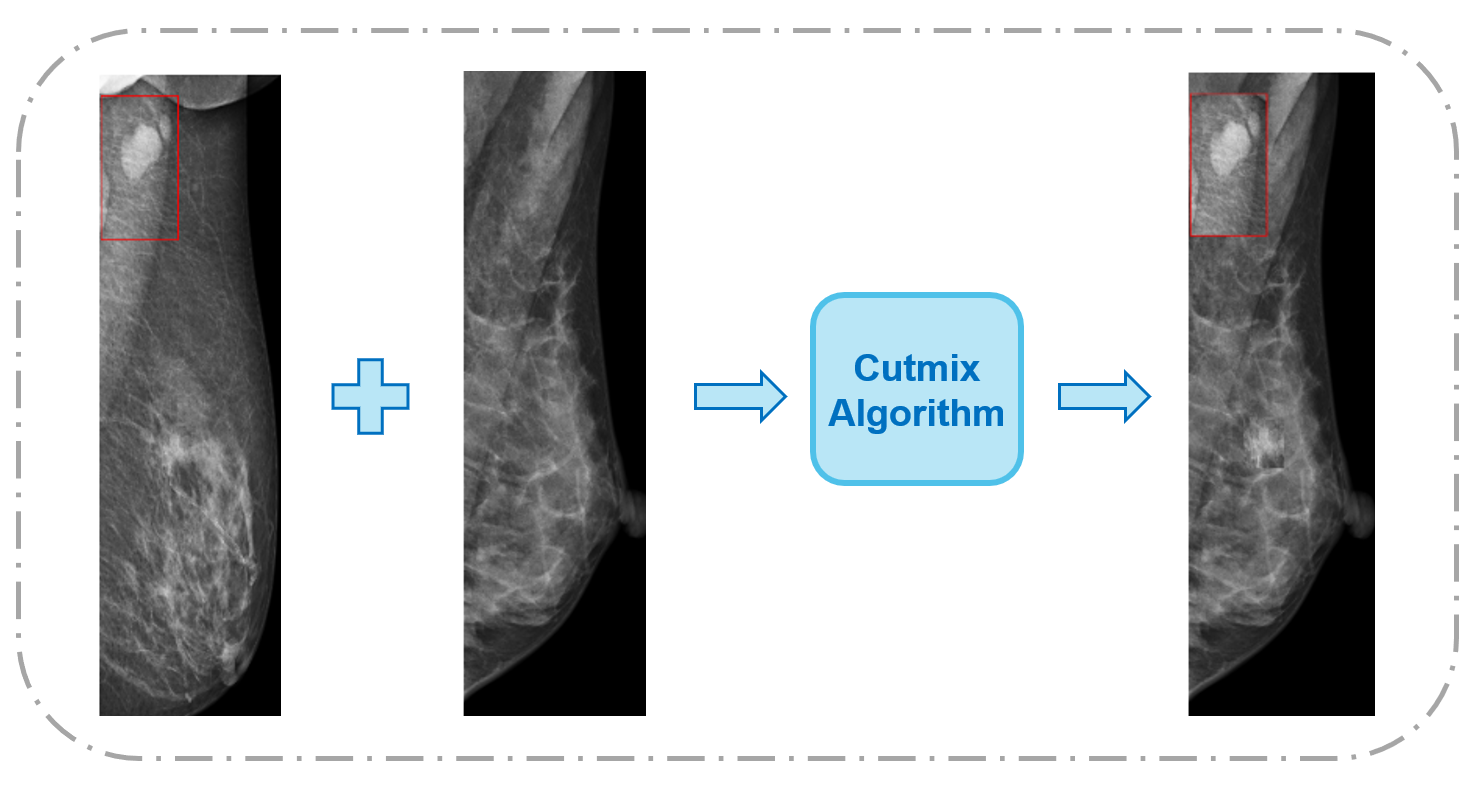}
         \caption{CutMix algorithm~\cite{cutmix}}
         \label{fig:2a}
     \end{subfigure}
     \begin{subfigure}[b]{0.48\textwidth}
         \centering
         \includegraphics[scale=.27]{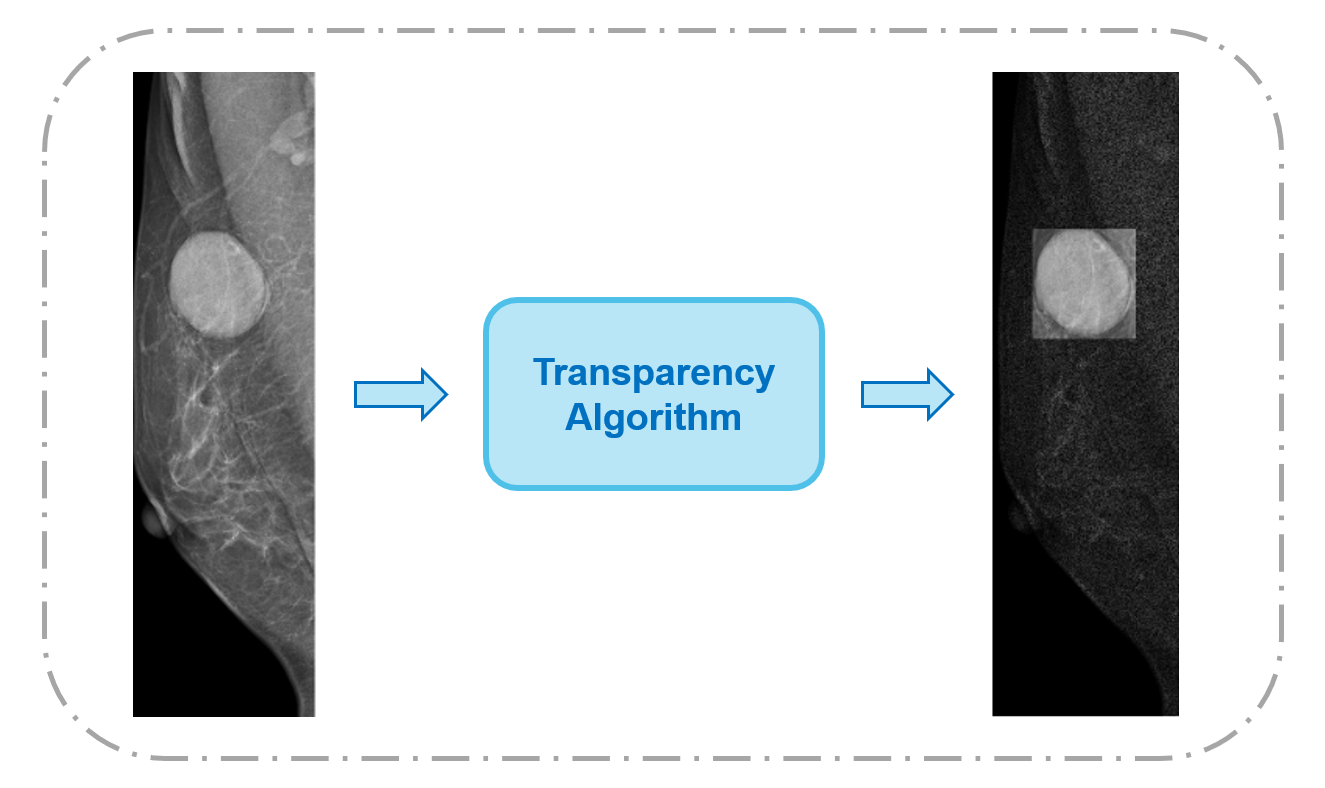}
         \caption{Transparency algorithm (\textbf{ours})}
         \label{fig:2b}
     \end{subfigure}
        \caption{\small{Illustration of the CutMix augmentation algorithm and the proposed Transparency algorithm.}}
        \label{fig:algorithms}
\end{figure*}
\\ \subsubsection{CutMix algorithm}
\label{sssec:subsub312}
The CutMix augmentation strategy~\cite{cutmix} was applied to mammograms to increase the number of unusual samples. The key idea behind the CutMix algorithm is to create a new pattern by merging the interpolation of both images and two labels. In this study, we simply moved the lesion region from high-risk scans to low-risk ones by changing pixel values. We denote that $x \in R^{W \times H \times C}$ and $y$ correspond to a mammography sample and its label where $W, H, C$ are width, height, and channels of this sample, respectively. First, we extract the lesion area from image A $(x_A, y_A )$ with abnormal bounding boxes $Box = (x_{min}, x_{max}, y_{min}, y_{max})$ and ground truth labels $y_A \in \{ \text{“BIRADS 3”}, \text{“BIRADS 4”}, \text{“BIRADS 5”}\}$. A mask $M \in \{0 ;1\}^{W \times H}$ is created by reducing the background pixel value to 0 and keeping the box pixels to 1 as 
\begin{equation}
M_{ij} = \Big\{^ {1 \: ; \text { if } (x_{min} \leq i \leq x_{max}) \& ( y_{min} \leq j \leq y_{max})}_{0 \: ; \text{ if } (x_{min} > i \: | \: i > x_{max} \: |\:  y_{min} > j \: | \:j > y_{max}).}
\end{equation}
We then generate the new sample $(x', y')$ through the equation
\begin{equation}
\begin{split}
x' = M \odot x_A + (1 - M) \odot x_B,\\
y' = y_A,
\end{split}
\end{equation} 
where $\odot$ denotes element-wise multiplication, image B $(x_B, y_B)$ with $y_B \in \{ \text{“BIRADS 1”}, \text {“BIRADS 2”}\}$.
\\\subsubsection{Transparency algorithm}
\label{sssec:subsub312}
Due to the lack of data, especially BI-RADS 3, BI-RADS 4, BI-RADS 5 mammograms, an image augmentation strategy called ``Transparency Strategy" was proposed to generate more high-risk samples from original images. The minimum condition for implementing this algorithm is that the training dataset contains images with unusual bounding boxes. A new training sample $(x', y')$ is created by transforming pixel values from the original sample $(x,y)$. The transformation operation will be described in detail below
\begin{equation}
\begin{split}
x' = M \odot x, \\
y' = y.
\end{split}
\end{equation} 
This operator is a label-preserving transformation, BI-RADS of the new sample is the original label. The sample increment algorithm artificially increases the training dataset size by preserving the pixels at the lesion and reducing the background pixel values of the bounding box image. For abnormal images, $Box = (x_{min}, x_{max}, y_{min}, y_{max})$ indicates the bounding box location. In the mask, $M \in \{\alpha ;1\}^{W \times H}$, $\alpha$ is a random number that ranges from $0.1$ to $0.9$, and $1$ is the value of pixels that is inside the lesion bounding box. The range of alpha was chosen empirically. Then the mask has been created with the following formulas
\begin{equation}
M_{ij} = \Big\{^ {1 \: ; \text { if } (x_{min} \leq i \leq x_{max}) \& ( y_{min} \leq j \leq y_{max})}_{\alpha \: ; \text{ if } (x_{min} > i \: | \: i > x_{max} \: |\:  y_{min} > j \: | \:j > y_{max})}.
\end{equation}
Before putting the data into the models, we define a data loader that loads the data and generates more samples based on the Transparency algorithm. In comparing several data augmentation techniques that use information from abnormal bounding boxes, such as CutMix~\cite{cutmix}, Mixup~\cite{mixup}, Cutout~\cite{cutout} the advantages of the proposed approach are using the whole breast image, utilizing informative lesion location, and adjusting the weight of the region needed to focus. See Figure~\ref{fig:algorithms} for more details. 
\\[-0.75cm]\subsection{Deep Learning BI-RADS Classification}
\label{ssec:subhead32}
We implemented a state-of-the-art deep learning model to classify BI-RADS on mammograms in this work. Our network consists of two main components: (i) an extraction feature based on the Efficientnet-B2~\cite{tan2019efficientnet} architecture, the output of which is a feature representation for each sample image, and (ii) a fully connected layer as a classifier to predict results from computed representations. We refer the readers to Tan \textit{et al.} ~\cite{tan2019efficientnet} for more detail about the network architecture.   
\section{Experiments}
\label{sec:ExperimentResult}
\subsection{Datasets and experimental setup}
\label{ssec:subhead50}
We evaluate the effectiveness of the proposed approach on three datasets: our private dataset, VinDr-Mammo~\cite{nguyen2022vindr}, and MIAS~\cite{mias}.
\\[0.3cm] \textbf{Our private dataset} was collected from Hospital 108 (H108) and Hanoi Medical University Hospital (HMUH) from 2018 to 2020. The dataset includes 36,614 screening mammogram images that come with their BI-RADS classification; each image was annotated by a team of three radiologists specializing in breast imaging for global labels (BI-RADS 1$\div$5). These images were divided into three groups by the global label stratification method~\cite{stratify}: training set (25,373), validation set (5,398), and test set (5,393). In the dataset, 2,503 images were remarked on three local labels (lesions), including Discrete Mass, Spiculated Mass, and Stellate Mass with bounding boxes. The number of lesions on BI-RADS 3 $\&$ 4 $\&$ 5 occupies almost a total of mass images that contains local labels. Descriptions of three sets on our internal dataset are provided in Table~\ref{tab:1}. \\[0.3cm]
\textbf{VinDr-Mammo dataset} is a large-scale benchmark full-field digital mammography dataset consisting of 5,000 four-view exams with breast-level BI-RADS and findings annotations. This is currently the largest public dataset with 20,000 scans providing BI-RADS assessment and suspicious/probably benign findings. Mammography images were acquired retrospectively from  H108 and HMUH. We divided this dataset by the given stratification method in~\cite{stratify}. Therefore, one-fifth (4,000 images) of the VinDr-Mammo dataset was used for testing and the rest part (16,000 images) of dataset was used for training.\\
\setlength\parindent{0pt} \textbf{MIAS dataset} was collected from the United Kingdom National Breast Screening Program and then labeled by specialists in 1994. This dataset contains 322 images that were divided into three classes: 209 images for normal, 61 images for benign, and 52 images for malignant class. We randomly stratified the dataset into training and test sets with a ratio of 0.8/0.2. As a result, there are 265 images used for training and 67 images for testing the classification algorithms.
\begin{table*}[ht!]
\centering
\scriptsize{
    \caption{\textbf{Description of the private dataset used for model development and validation.}}
    \label{tab:1}
    \begin{tabular}{||c||c|c|c|c|c||c|c|c|c|c||c|c|c|c|c||}
    \hline
    \textbf{Data}    & \multicolumn{5}{c||}{\textbf{Training set}}       & \multicolumn{5}{c||}{\textbf{Validation set}}    & \multicolumn{5}{c||}{\textbf{Test set}}       \\
    \hline
    \diaghead{\theadfont Diag ColumnmnHead}{\textbf{Lesions}}{\textbf{BI-RADS}} & \textbf{1}   & \textbf{2}   & \textbf{3}  & \textbf{4}  & \textbf{5}  & \textbf{1}   & \textbf{2}   & \textbf{3}  & \textbf{4}  & \textbf{5} & \textbf{1}   & \textbf{2}   & \textbf{3}  & \textbf{4}  & \textbf{5} \\
    \hline
   Discrete mass      & 2 & 31 & 1,077 & 779 & 153 & 0   & 3   & 245   & 193   & 25 & 2   & 7  & 237   & 168   & 35 \\
    Spiculated mass     & 0 & 1 & 1 & 49 & 226  & 0   & 0   & 0   & 11   & 41 & 0   & 0   & 0   & 18   & 48 \\
   Stellate mass       & 0   & 0   & 1   & 70   & 95  & 0    & 0    & 0    & 8    & 21   & 0    & 0    & 3    & 17    & 21   \\
    \textbf{Total mass images}       & 2   & 13   & 901   & 571   & 222  & 0    & 2    & 217    & 134    & 38   & 2    & 4    & 201    & 138    & 58   \\
    \textbf{Total images}   & 16,203 & 5,435 & 1,699 & 1,514 & 522 & 3,385 & 1,208 & 372 & 325 & 108 & 3,376 & 1,191 & 369 & 338 & 119 \\
    \hline
    \end{tabular}}
\end{table*}

\subsection{Training Methodology \& Evaluation Metrics }
\label{ssec:subhead51}
Our experiments were built on PyTorch and using a PC with an Nvidia GTX 1080 GPU. We trained the feature extractors using SGD optimizer~\cite{ruder2016overview} with a momentum of 0.9 and cosine annealing learning rate~\cite{CALR}. The cross-entropy function was used to calculate the error. For model evaluation, we used \textit{F1}-score on the 5-class BI-RADS level. \textit{F1}-score is the harmonic mean of precision and recall. For multi-class problems, the \textit{F1}-score macro, which is defined as the mean of class-wise \textit{F1}-scores could be used. In our experiments, the results are appraised on image-level for BI-RADS classification. The classification models for different augmentation methods are trained with the same network architecture (EfficientNet-B2~\cite{tan2019efficientnet}) and a fixed image size of $\textit{1024} \times \textit{768}$. The number of epochs was set to 50, and the training process stopped in case there was no improvement in the \textit{F1}-score of the validation set after 15 consecutive epochs by an early stopping callback. The performance of different techniques is assessed on the test set with the same evaluation metrics and network architecture.
\\[-0.75cm]\subsection{Experimental Results}
\label{ssec:subhead52}
This section reports the effectiveness of the proposed Transparency approach and compares its performance with two other data augmentation techniques: baseline and CutMix~\cite{cutmix}. We observed that our method consistently outperforms the baseline and the state-of-the-art method by a large margin in different settings.
\\\subsubsection{Comparison with the baseline}
The baseline model is trained on our private dataset, VinDr-Mammo, and MIAS datasets, in which all images are cropped with the breast detector and flipped before fitting into the model. Compared to the baseline, the proposed transparency approach achieved better classification performance of about 4.3$\%$ higher in F1-scores on the private dataset, 6.9$\%$ and 2.7$\%$ higher on the VinDr-Mammo and MIAS dataset, respectively. Table~\ref{tab:2} reports the experimental result on these three datasets.  
\\\subsubsection{Comparison with a state-of-the-art technique}
Experiments on the private dataset and VinDr-Mammo showed that both CutMix~\cite{cutmix} and the proposed transparency technique considerably enhance the performance of the classifier. In particular, the proposed method performed significantly better than the CutMix~\cite{cutmix} algorithm for all classes on the VinDr-Mammo dataset, and for three classes such as BI-RADS 1, BI-RADS 2, and BI-RADS 3 on our in-house dataset. In particular, the Macro-F1 of the proposed method surpassed the CutMix algorithm by around 1\% on the private dataset and by 6.5\% on VinDr-Mammo. The experimental result on these two datasets was illustrated in Table~\ref{tab:3}.



\begin{table}[t!]
\centering
\scriptsize{
    \caption{Quantitative results (F1-score) of baseline and Transparency technique on the private dataset, VinDr-Mammo, and MIAS datasets. The highest score are highlighted in \textcolor{red}{\textbf{bold}.}}
    \label{tab:2}
    \begin{tabular}{||c|c|c|c||}
    \hline
     \textbf{Dataset} & \textbf{Class} &\textbf{ Baseline} & \textbf{Transparency (ours)} \\
    \hline
    \multirow{6}{*}{\textbf{Private dataset}} & BI-RADS 1 & 0.85 & \textcolor{red}{\textbf{0.86}} \\
                                     & BI-RADS 2 & 0.59 & \textcolor{red}{\textbf{0.61}} \\
                                     & BI-RADS 3 & 0.21 & \textcolor{red}{\textbf{0.31}} \\
                                     & BI-RADS 4 & 0.52 & \textcolor{red}{\textbf{0.57}} \\
                                     & BI-RADS 5 & 0.60 & \textcolor{red}{\textbf{0.63}} \\
                                     & Macro - F1 & 0.552 & \textcolor{red}{\textbf{0.595}} \\
    \hline
    \multirow{6}{*}{\textbf{VinDr-Mammo}} & BI-RADS 1 & \textcolor{red}{\textbf{0.89}} & \textcolor{red}{\textbf{0.89}} \\
                                     & BI-RADS 2 & 0.59 & \textcolor{red}{\textbf{0.61}} \\
                                     & BI-RADS 3 & 0.55 & \textcolor{red}{\textbf{0.68}} \\
                                     & BI-RADS 4 & \textcolor{red}{\textbf{0.51}} & 0.50\\
                                     & BI-RADS 5 & 0.50 & \textcolor{red}{\textbf{0.69}} \\
                                     & Macro - F1 & 0.607 & \textcolor{red}{\textbf{0.676}} \\

    \hline
    \multirow{4}{*}{\textbf{MIAS}} & Normal & 0.70 & \textcolor{red}{\textbf{0.71}} \\
                                     & Benign & \textcolor{red}{\textbf{0.32}} & 0.28 \\
                                     & Malignant & 0.27 & \textcolor{red}{\textbf{0.37}}\\
                                     & Macro-F1 & 0.428 & \textcolor{red}{\textbf{0.455}} \\
    \hline
    \end{tabular}}
\end{table}

\begin{table}[t!]
\centering
\scriptsize{
    \caption{Quantitative results (F1-score) using CutMix and Transparency technique on the private dataset and VinDr-Mammo dataset. The highest score are highlighted in \textcolor{red}{\textbf{bold}}.}
    \label{tab:3}
    \begin{tabular}{||c|c|c|c|c||}
    \hline
    {\textbf{Class}} & \multicolumn{2}{c|}{\textbf{Private dataset} } & \multicolumn{2}{c||}{\textbf{VinDr-Mammo}} \\
    \cline{2-5}
    & CutMix & Transparency & CutMix & Transparency \\
    && (ours) & & (ours)\\
    \hline
    {\textbf{BI-RADS 1}} & 0.85 & \textcolor{red}{\textbf{0.86}} & \textcolor{red}{\textbf{0.89}} & \textcolor{red}{\textbf{0.89}} \\
    {\textbf{BI-RADS 2}} & 0.59 & \textcolor{red}{\textbf{0.61}} & 0.59 & \textcolor{red}{\textbf{0.61}} \\
    {\textbf{BI-RADS 3}} & 0.25 & \textcolor{red}{\textbf{0.31}} & 0.53 & \textcolor{red}{\textbf{0.68}} \\
    {\textbf{BI-RADS 4}} & \textcolor{red}{\textbf{0.57}} & \textcolor{red}{\textbf{0.57}} & 0.48 & \textcolor{red}{\textbf{0.50}}\\
    {\textbf{BI-RADS 5}} & \textcolor{red}{\textbf{0.68}} & 0.63 & 0.56 & \textcolor{red}{\textbf{0.69}}  \\
    \hline
    {\textbf{Macro-F1}} & 0.589 & \textcolor{red}{\textbf{0.595}} & 0.611 & \textcolor{red}{\textbf{0.676}}  \\
    \hline
    \end{tabular}}
\end{table}

\section{Conclusion}
\label{sec:conclude}
Resolving data imbalance is one of the most challenging problems in machine learning, especially in medical image analysis, where anomalous patterns are rare and hard to collect. This study introduced a Transparency strategy-based technique that creates diseased samples by adjusting the pixel values. Experimentally, the proposed approach shows strong evidence that it could improve the BI-RADS classification task on mammogram exams. Our approach is simple and can be applied to various tasks in medical imaging, especially for lesion detection and classification problems.\\
\\
\noindent \textbf{Compliance with Ethical Standards}.
Our work follows all applicable ethical research standards and laws. The study has been reviewed and approved by the hospital's institutional review board (IRB). The need for obtaining informed patient consent was waived because this work did not impact clinical care.
\\
\noindent \textbf{Acknowledgements}.
This study was supported by Smart Health Center, VinBigData JSC. We would like to acknowledge Hanoi Medical University Hospital for providing access to their image databases. In particular, we thank all of our radiologists who participated in this project.

\bibliographystyle{abbrv}
\bibliography{refs}
\end{document}